\documentclass{article}
\usepackage{spconf,amsmath,graphicx}

\usepackage{amssymb}
\newcommand{\bv}[1]{\mathbf{#1}}
\DeclareMathOperator*{\argmin}{argmin}

\newcommand{\Tref}[1]{Table~\ref{#1}}
\newcommand{\Eref}[1]{Eq.~(\ref{#1})}
\newcommand{\Fref}[1]{Fig.~\ref{#1}}

\newcommand{\Sref}[1]{Sec.~\ref{#1}}
\usepackage{booktabs}
\usepackage{multirow}

\usepackage{subcaption}

\usepackage{url}

\usepackage{eso-pic,xcolor}
\AddToShipoutPictureBG*{%
  \put(\LenToUnit{.1\paperwidth},\LenToUnit{4em})%
    {\parbox[3pt]{49em}{%
      \footnotesize\copyright 2022 IEEE. Personal use of this material is permitted. Permission from IEEE must be obtained for all other uses, in any current or future media, including reprinting/republishing this material for advertising or promotional purposes, creating new collective works, for resale or redistribution to servers or lists, or reuse of any copyrighted component of this work in other works.
      }%
    }%
}

\title{ARM 4-bit PQ: SIMD-based Acceleration for Approximate Nearest Neighbor Search on ARM}
\name{Yusuke Matsui$^{\star}$ \qquad Yoshiki Imaizumi$^{\dagger}$ \qquad Naoya Miyamoto$^{\dagger}$ \qquad Naoki Yoshifuji$^{\dagger}$}
\address{$^{\star}$ The University of Tokyo \qquad $^{\dagger}$ Fixstars Corporation}

\begin{document}
%
\maketitle
\begin{abstract}
We accelerate the 4-bit product quantization (PQ) on the ARM architecture. Notably, the drastic performance of the conventional 4-bit PQ strongly relies on x64-specific SIMD register, such as AVX2; hence, we cannot yet achieve such good performance on ARM. To fill this gap, we first bundle two 128-bit registers as one 256-bit component. We then apply shuffle operations for each using the ARM-specific NEON instruction. By making this simple but critical modification, we achieve a dramatic speedup for the 4-bit PQ on an ARM architecture. Experiments show that the proposed method consistently achieves a 10x improvement over the naive PQ with the same accuracy.
\end{abstract}

\section{Introduction}
\label{sec:intro}

The approximate nearest neighbor search is a fundamental technique in many fields of computer science, such as signal processing, computer vision, and natural language processing. In this paper, we focus on the product quantization (PQ) family of nearest neighbor methods~\cite{tpami_jegou2011, ite_matsui2018, tpami_ge2014, icml_zhang2014, cvpr_babenko2014, eccv_martinez2016, eccv_martinez2018}. When data are too large to be stored directly in memory, compressing them using PQ is the de facto standard for data handling. One can then perform the nearest neighbor search using a table lookup on the compressed data. One of the fastest methods is 4-bit PQ~\cite{vldb_andre2015, icmr_andre2017}, which can search $10^6$ vectors in less than 1 ms. Notably, previous approaches were usually evaluated on the x86 architecture.

Meanwhile, searching using architectures other than x86 has gained significant attention recently. For example, ARM architecture has gained rapid popularity, from its applications to advanced consumer computers (e.g., Apple M1) to high-performance cloud resources (e.g., the Amazon Web Services (AWS) Graviton2). PQ-based methods often rely on SIMD instructions specified for x86, such as AVX2 and AVX512. Therefore, there is a great demand for the fast realization of PQ-based methods for ARM.

The technical problem is that the size of the SIMD registers for ARM is usually smaller than that of x86, implying that we cannot make full use of SIMD-based acceleration. 256-bit SIMD registers (i.e., AVX2) are typically available for recent x86 architectures, whereas ARM typically use registers with 128-bit or less (i.e., the NEON family).

We propose a simple but drastically efficient approach for fast 4-bit PQ searching on ARM. The main idea is to handle two 128-bit SIMD registers as if they were a single 256-bit register. In doing so, we can apply a shuffle operation of NEON instructions instead of the x86-specific AVX2. Our contributions are as follows:
\begin{itemize}
\item The proposed 4-bit PQ for ARM is 10x faster than the naive ARM implementation when applied to the SIFT1M and Deep1M datasets.
\item The proposed approach, combined with inverted indexing and hierarchical navigable small-world (HNSW) graphs, achieves an efficient search on Deep1B data (less than 1 ms/query). We believe this result is the current record for the fastest nearest neighbor search on the ARM architecture.
\item The proposed method has already been implemented in the Faiss library and is available freely~\cite{web_faiss_simdlib_neon}.
\end{itemize}

\begin{figure*}
\centering
\subcaptionbox{ADC\label{fig:comp1} ($K=256$)}[0.26\linewidth]{
    \includegraphics[width=1.0\linewidth]{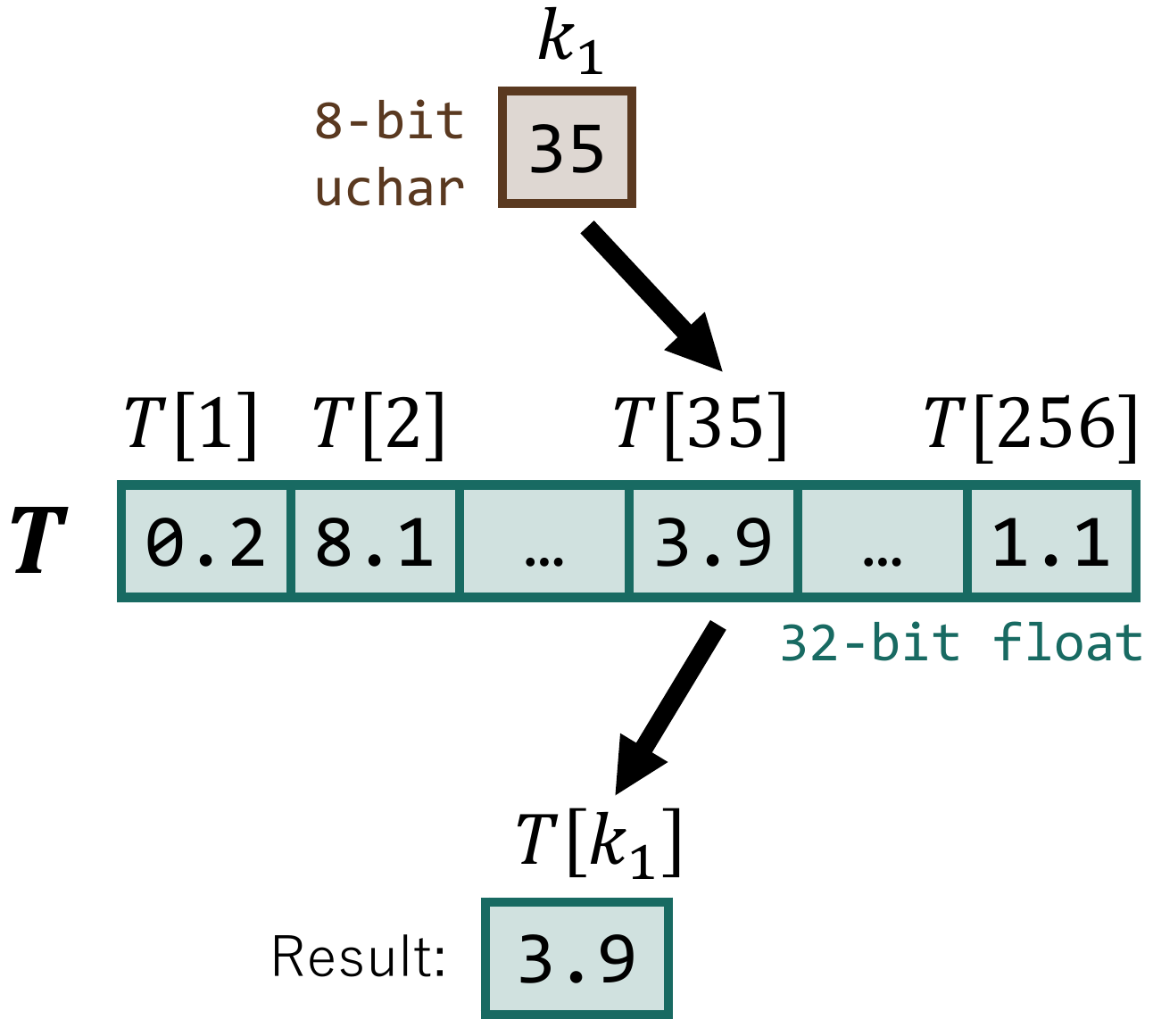}
}
\quad
\subcaptionbox{ADC with 128-bit SIMD register \label{fig:comp2}}[0.26\linewidth]{
    \includegraphics[width=1.0\linewidth]{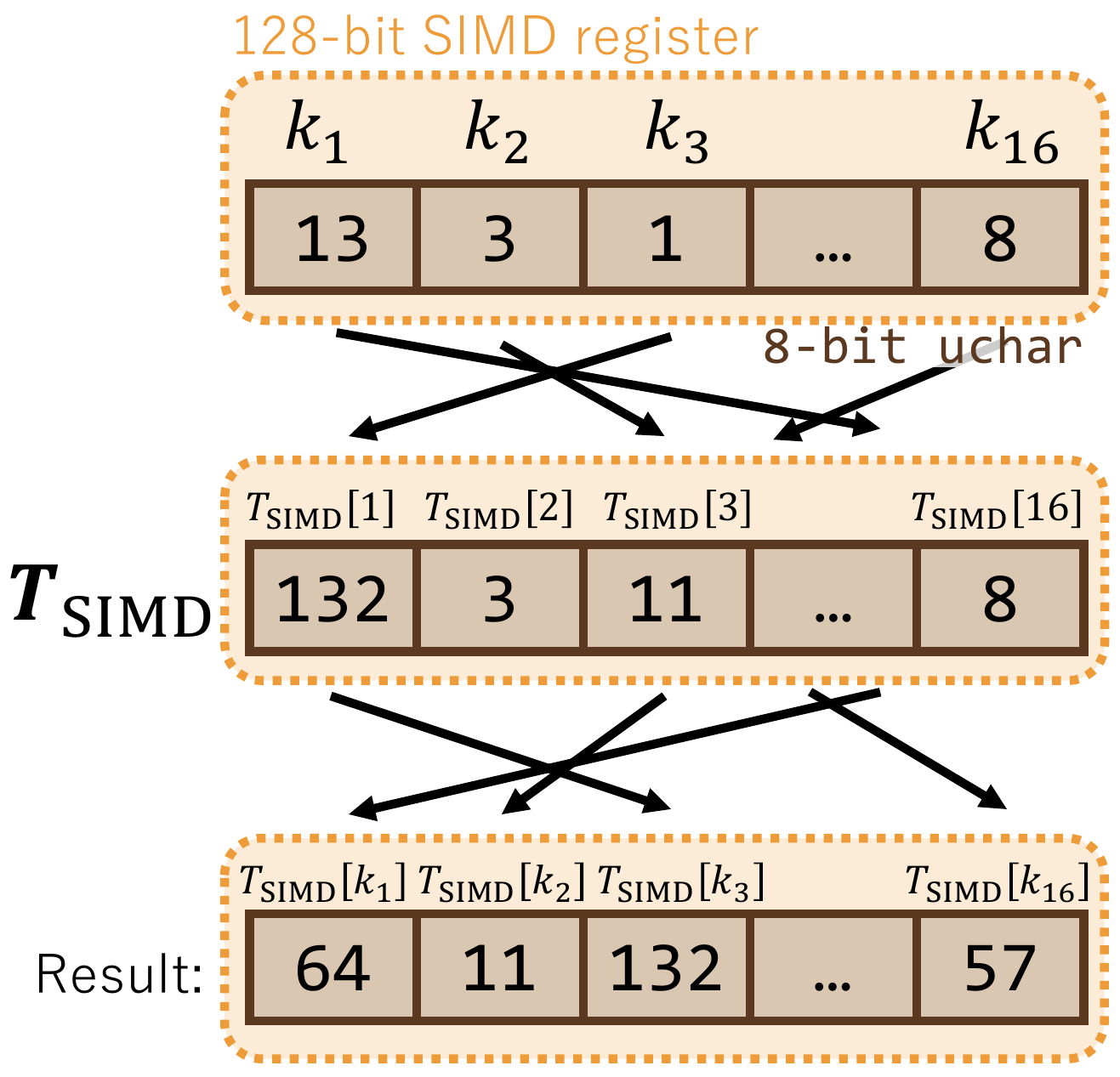}
}
\quad
\subcaptionbox{Proposed: ADC with 128x2-bit SIMD registers on ARM\label{fig:comp3}}[0.43\linewidth]{
    \includegraphics[width=1.0\linewidth]{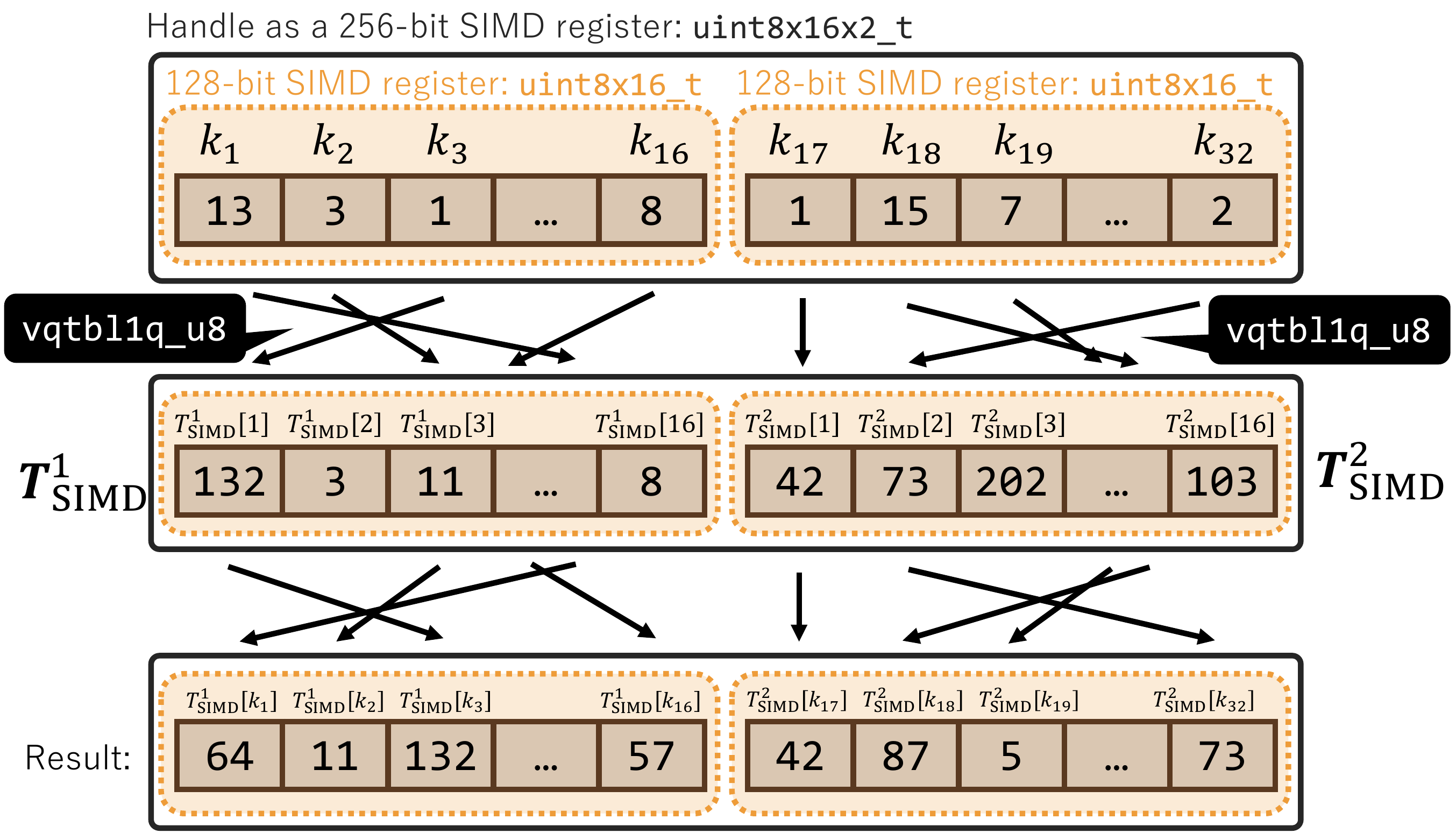}
}
\vspace{-5mm}
\caption{Comparison of asymmetric distance computation (ADC) lookup operations}\label{fig:comp}
\end{figure*}

\section{Preliminarily}
We first describe the problem setting and introduce the most relevant approach.
Let $\{ \bv{x}_n \}_{n=1}^N$ be $N$ $D$-dimensional database vectors, where $\bv{x}_n \in \mathbb{R}^D$.
Given a query vector, $\bv{q} \in \mathbb{R}^D$, our task is to find the database vector that minimizes the (squared) Euclidean distance: $\Vert \bv{q} - \bv{x}_n \Vert_2^2$. We can find this vector using a simple linear scan, but doing so presents two problems. First, we need at least $32ND$ bits of memory space to store the database vectors (using 32-bit floats). This cost is prohibitive if the database is large, such as when $N=10^9$ and $D=100$. Second, the computational complexity is $O(ND)$, which is practically slow.

\textbf{Product Quantization}~\cite{tpami_jegou2011}: To solve the memory issue, the de facto approach is to compress each database vector by PQ~\cite{tpami_jegou2011}, where a vector is split into sub-vectors, each quantized by traditional vector quantization (VQ)~\cite{assp_gray1984}. As PQ is a straightforward extension of VQ, we discuss VQ in this section for simplicity. We then extend the discussion to PQ at the end of \Sref{sec:proposed}.

Given an input vector, $\bv{x}\in\mathbb{R}^D$, let us define a vector quantizer, $Q:\mathbb{R}^D \to \{1, \dots, K\}$, as follows:
\begin{equation}
    Q(\mathbf{x}) = \argmin_{k\in \{1, \dots, K\}} \Vert \bv{x} - \bv{c}_k\Vert_2^2.
\end{equation}
Here, $\{\bv{c}_k\}_{k=1}^K \subset \mathbb{R}^D$ provides codewords, which are typically created by running k-means clustering over the training dataset. Note that $K$ is the number of codewords.
With $Q$, a $D$-dimensional vector is quantized into a short code (integer) $k$,
which can be represented by only $\log_2 K$ bits.
Here, $K$ is usually set to 256, so that each code takes $\log_2 256 = 8$ bits = $1$ B (unsigned char).
Given $k$, we can lossy-reconstruct the original vector, $\bv{x}$, by fetching the codeword, $\bv{c}_k$.

We apply $Q$ for each database vector, $\bv{x}_n$, which results in the quantized codes, $\{k_n \}_{n=1}^N$, where $k_n=Q(\bv{x}_n) \in \{1, 2, \dots, K \}$. Here, we discard the original database vectors, $\{\bv{x}_n\}_{n=1}^N$, and maintain only $\{k_n \}_{n=1}^N$.
Compared with the original vectors, which require $32ND$ bits, the quantized codes require only $N\log_2 K$ bits.

Given $\bv{q}$, our task is to find a similar item from the quantized codes, $\{k_n\}_{n=1}^N$.
A straightforward approach is to reconstruct all vectors,
$\bv{c}_{k_1}, \bv{c}_{k_2}, \dots, \bv{c}_{k_N}$, and run a linear scan, but this requires considerable time and memory.
Alternatively, we can obtain the same result via an asymmetric distance computation (ADC)~\cite{tpami_jegou2011}.
Let us define a $K$-dimensional vector, $\bv{T} \in \mathbb{R}^K$, whose $k$-th element is defined as follows:
\begin{equation}
    T[k] = \Vert \bv{q} - \bv{c}_k \Vert_2^2.
\end{equation}
We create $\bv{T}$ only once when given $\bv{q}$. To approximate the distance between $\bv{q}$ and $\bv{x}_n$, we look up $\bv{T}$ by $k_n$ as the key:
\begin{equation}
    \Vert \bv{q} - \bv{x}_n \Vert_2^2 \sim \Vert \bv{q} - \bv{c}_{k_n} \Vert_2^2 = T[k_n].
    \label{eq:approx_lut}
\end{equation}
This is a fast $O(1)$ operation.
\Fref{fig:comp1} visualizes the operation.

\textbf{4-bit PQ}~\cite{vldb_andre2015, icmr_andre2017}: 
Although the table lookup discussed above is fast, it is not ``extremely'' fast because (1) we must use the main memory for the lookup, and (2) the entire operation lacks concurrency. A faster approach is to look up the element from the SIMD register~\cite{vldb_andre2015, icmr_andre2017}, which is a special hardware that can efficiently perform the same arithmetic operation on multiple elements. The problem is that the SIMD register is small (typically 128 or 256 bits). Thus, we must modify the algorithm to fully employ it. The 4-bit PQ achieves an approximate but fast PQ via the SIMD register.

We next describe the 4-bit PQ.
Recall that $\bv{T}$ is a $K$-dimensional vector. Thus, $\bv{T}$ is represented by a $K$-length array with floating points ($32K$ bits total). Under the typical setting, $K=256$, the total memory cost of $\bv{T}$ is 8,192 bits; hence, we cannot load it using the SIMD register. Hence, we apply a drastic approximation:
\begin{itemize}
    \item Set $K=16$ so that each code takes only $\log_2 16 = 4$ bits. This explains the ``4-bit’’ aspect of PQ.
    \item Apply a scalar quantization for each element in $\mathbf{T}$, so that each element is represented by an 8-bit unsigned char.
\end{itemize}
By this, we obtain a new look-up table, $\mathbf{T}_\mathrm{SIMD} \in \{1, \dots, 256\}^{16}$.
Using this table, we can replace the \Eref{eq:approx_lut} as follows:
\begin{equation}
\Vert \bv{q} - \bv{x}_n \Vert_2^2 \sim \Vert \bv{q} - \bv{c}_{k_n} \Vert_2^2 \sim f(T_\mathrm{SIMD}[k_n]),
\end{equation}
where $f$ is the reconstruction of an unsigned char to float, which is trivial.
This table is represented by an array of 16 unsigned chars, which is 128 bits total and can be loaded onto the SIMD register.
Because each code, $k_n$, is represented by only 4 bits, we can load 16 elements (e.g., $k_1$ to $k_{16}$) at once. Using the shuffle operation, we can run the table lookup inside the SIMD register in parallel, as shown in \Fref{fig:comp2}. Then, we store the result in another SIMD register.
This approach is fast because the SIMD-register lookup is much faster than a memory lookup, and we can process 16 elements at once.

Similar SIMD-based formulations have been proposed (Bolt~\cite{kdd_blalock2017}, ScaNN~\cite{icml_guo2020}, and MADNESS~\cite{icml_blalock2021}). The Faiss library~\cite{tpami_johnson2021} also supports the SIMD-based approach\footnote{In Faiss, the 4-bit PQ algorithm is referred to by the class name \texttt{PQFastScan}, from the original paper's title\cite{vldb_andre2015, icmr_andre2017}.}. For the latest x86 architectures, the AVX512-based method has also been proposed~\cite{tpami_andre2021}.

\section{Proposed approach}
\label{sec:proposed}

We next describe our approach for achieving a 4-bit PQ for the ARM architecture. We select the Faiss implementation for our baseline as it is the de facto library for approximate nearest neighbor (ANN) searching.
The 256-bit SIMD register (the \texttt{\_\_m256i} data type for AVX2) is used for 4-bit PQ Faiss searching on x86 architectures.
Using a 256-bit register, we can look up \Eref{eq:approx_lut} twice with a single SIMD operation (i.e., the 256-bit shuffle operation; \texttt{\_mm256\_shuffle\_epi8}). Thus, we can operate 32 elements at once.
This choice is natural for the x86 architecture because all applicable modern computers have 256-bit SIMD registers.

However, the problem is that the ARM architecture usually does not have 256-bit registers. For example, only 64- and 128-bit SIMD registers are available for ARMv7 and ARMv8, respectively.
Moreover, ARM does not offer a 256-bit lookup operation.

Our approach is simple; we concatenate two 128-bit SIMD registers and use them as if it is a single 256-bit register (\texttt{uint8x16x2\_t}).
We run a 128-bit lookup operation (\texttt{vqtbl1q\_u8}) twice: one for the first 128 bits of the register and the other for the last 128.
These operations achieve the same result as the \texttt{\_mm256\_shuffle\_epi8} of AVX2.

We visualize our approach in \Fref{fig:comp3}. Let us define the two tables as $\bv{T}_\mathrm{SIMD}^1, \bv{T}_\mathrm{SIMD}^2 \in \{1, \dots, 256\}^{16}$. Each table takes 128 bits, as discussed.
We stack these two registers and handle the whole as a 256-bit component.
To perform a lookup operation, the system receives 32 8-bit unsigned chars as \texttt{uint8x16x2\_t} (e.g., $k_1, \dots, k_{32}$). We run the 128-bit shuffle operation, \texttt{vqtbl1q\_u8}, twice as follows:
\begin{itemize}
    \item For the first 128 bits ($k_1, \dots, k_{16}$), we run the shuffle with $\bv{T}_\mathrm{SIMD}^1$.
    \item For the remaining 128 bits ($k_{17}, \dots, k_{32}$), we run the shuffle with $\bv{T}_\mathrm{SIMD}^2$.
\end{itemize}
The result is also stored in the \texttt{uint8x16x2\_t} registers.
With the proposed simple approach, we emulate a 256-bit 4-bit PQ operation using two 128-bit SIMD registers.

\textbf{From VQ to PQ}:
The above discussion is regarding a VQ that directly quantizes an input vector. PQ, however, splits the input vector into $M$ $D/M$-dimensional sub-vectors. VQ is then applied for each sub-vector.
With this adaptation, the memory cost for each vector becomes $M \log_2 K$ bits. Thus, for a typical $K=256$ setting, the cost is $8M$ bits. For a 4-bit PQ with $K=16$, the cost is $4M$ bits.
Note that we must carefully maintain the code layout~\cite{vldb_andre2015, icmr_andre2017}.
Optimizing the layout is also a central topic of 4-bit PQ, but it is outside the scope of this paper.

\textbf{Implementation detail}:
From a data-structure perspective, the provision of an interface that is equivalent for both x86 and ARM is essential. In this research, we designed a transparent interface for ARM by appropriately reinterpreting the x86-optimized base class for 256-bit registers in Faiss. We also implemented auxiliary instructions that are only present in AVX2 but not in ARM, including \texttt{\_mm256\_movemask\_epi8}. Please refer to our source code for more details~\cite{web_faiss_simdlib_neon}.

\section{Connection to existing approaches}
\label{sec:connection}

Next, we discuss the connection between 4-bit PQ and other ANN methods. Recall that 4-bit PQ is used on compressed data. Hence, 4-bit PQ is a building block of an ANN system.

\textbf{Inverted index}: 
To handle large-scale datasets, inverted indexing is widely used~\cite{tpami_jegou2011}.
The idea is to split the entire dataset ($\mathcal{X} = \{\mathbf{x}_n\}_{n=1}^N$)
into $n_\mathrm{list}$ smaller disjoint subsets: $\mathcal{X}_1, \dots, \mathcal{X}_{n_\mathrm{list}}$,
where $\mathcal{X}_i \cap \mathcal{X}_j = \emptyset$ for all $i, j$. Note that $\bigcup \mathcal{X}_i = \mathcal{X}$. 
Let $\boldsymbol{\mu}_i \in \mathbb{R}^D$ be a representative vector for $\mathcal{X}_i$. We then run the search as follows~\cite{ite_matsui2018}:
\begin{enumerate}
    \item For \textit{coarse quantization}, run the search between the query, $\mathbf{q}$, and the representative vectors, $\boldsymbol{\mu}_1, \dots, \boldsymbol{\mu}_{n_\mathrm{list}}$.
    \item Select the $n_\mathrm{probe}$ nearest representative vectors (i.e., the $n_\mathrm{probe}$ nearest subsets).
    \item For \textit{distance estimation}, run the second (fine) search for the selected subsets.
\end{enumerate}
This inverted index provides a general framework that we can use with any NN algorithm for coarse quantization.

We can also compress the data via PQ to handle large data in memory. For this case, we can achieve distance estimation via the ADC of the PQ.
Because 4-bit PQ is a special case of the usual PQ, we can use 4-bit PQ for inverted indexing.

\textbf{Graph-based approaches}
The current best algorithms for million-scale datasets are graph-based methods~\cite{icml_baranchuk2019,tpami_malkov2020,vldb_fu2019,icml_prokhorenkova2020,is_aumuller2020,tkde_wen2020}.
HNSW is an example~\cite{tpami_malkov2020} that is surprisingly fast and accurate, but it requires vast memory resources.

Our 4-bit PQ and HNSW are complementary. The purpose of a 4-bit PQ is to reduce the memory cost while providing an efficient distance estimation; hence, we can combine the HNSW and 4-bit PQ in the inverted index framework: (1) using HNSW for \textit{coarse quantization}, and (2) using 4-bit PQ for \textit{distance estimation}.
This combination of inverted index + HNSW + PQ is a common approach to handling large-scale data when they cannot be directly loaded into memory~\cite{icmr_tanaka2021, eccv_baranchuk2018}. We evaluate this hybrid approach in \Sref{sec:eval_large}.

\begin{figure*}
\centering
\subcaptionbox{Results on the SIFT1M dataset\label{fig:exp_comp_sift}}[0.48\linewidth]{
    \includegraphics[width=1.0\linewidth]{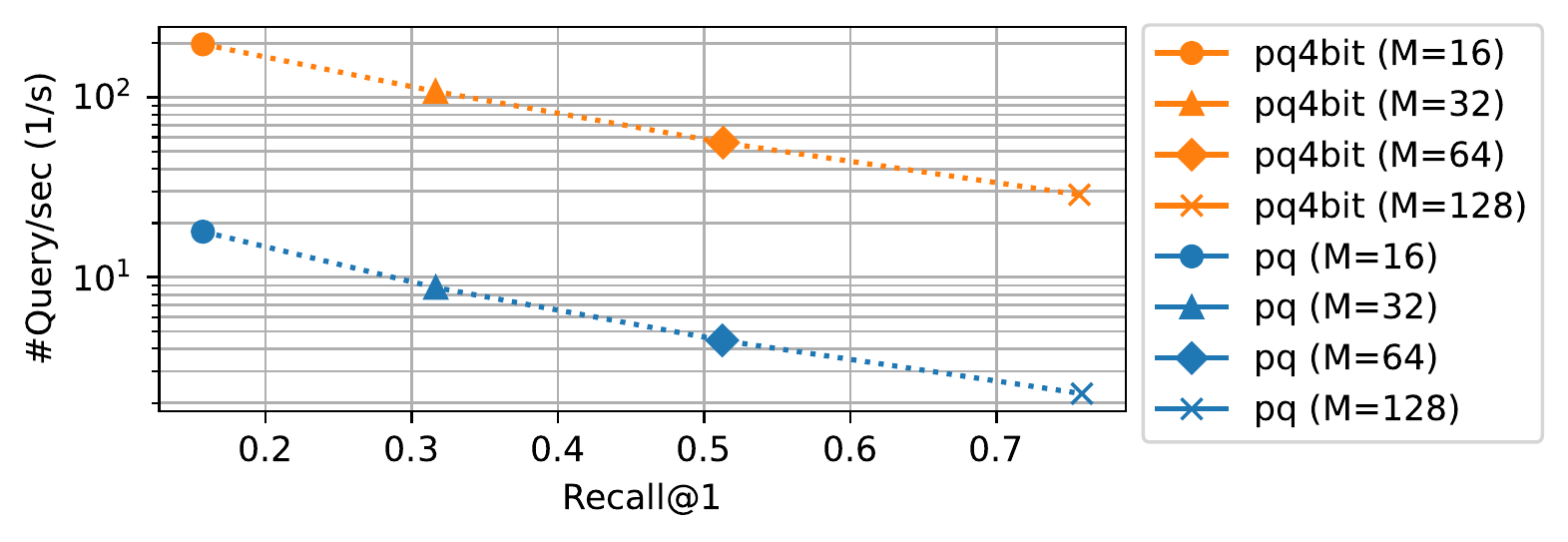}
}
\quad
\subcaptionbox{Results on the Deep1M dataset\label{fig:exp_comp_deep}}[0.48\linewidth]{
    \includegraphics[width=1.0\linewidth]{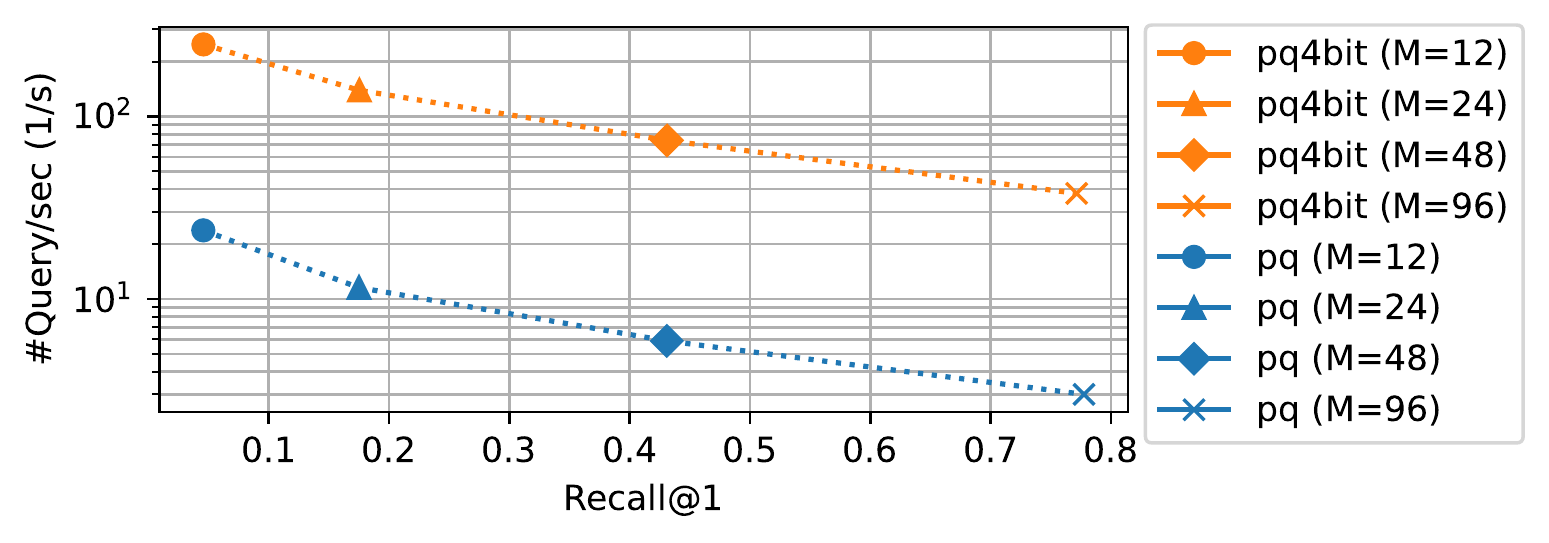}
}
\caption{Comparison between the proposed 4-bit product quantization (PQ) and standard PQ. Note that $K=16$ for all cases (i.e., each vector takes $M\log_2K=4M$ bits).}
\label{fig:exp_comp}
\end{figure*}

\section{Evaluation}
We performed all experiments on the AWS platform using a high-performance ARM computer: a \texttt{c6g.4xlarge} instance with a Graviton2 processor (16 vCPU and 32-GB RAM).
We implemented our method on the faiss codebase. 
We evaluated all methods with a single thread for fair comparison.
All results were the average of five trials.

We used the following datasets for our evaluation:
\begin{itemize}
    \item The Deep1B dataset~\cite{cvpr_babenko2016} consists of 96D deep features with $10^9$ base, $10^4$ query, and $10^6$ training vectors; we used the top $10^6$ vectors for the original training set.
    \item The Deep1M dataset is a subset of the Deep1B dataset (the top $10^6$ and the top $10^5$ vectors for the base and the training set, respectively).
    \item The SIFT1M dataset~\cite{icassp_jegou2011} consists of 128D features with $10^6$ base, $10^4$ query, and $10^5$ training vectors.
\end{itemize}

\subsection{Comparison with the original PQ}
\Fref{fig:exp_comp} shows the comparison between the original PQ and our 4-bit PQ for the SIFT1M and Deep1M datasets.
In these figures, the farther the point to the right , the more accurate the method; furthermore, the higher the point, the faster.

For both datasets, the two approaches achieved the same accuracy for the same $M$, whereas
\textbf{our 4-bit PQ always achieved 10x faster searches than the original PQ}.
This dramatic speedup was made possible by the proposed SIMD operation with the NEON instruction.

\subsection{Large-scale evaluation}
\label{sec:eval_large}
In this subsection, we show that our approach works well for the large-scale Deep1B dataset.
For billion-scale datasets, we can run neither linear scans nor HNSW, which handles the data directly without compression.
As discussed in \Sref{sec:connection}, we used the inverted index framework with HNSW and the 4-bit PQ.
The parameters of the space-division were set to $n_\mathrm{list}=30,000$ by following a well-known heuristic: $\sqrt{N}$.
We compressed each vector using 4-bit PQ with $K=16$ and $M=16$, resulting in 64 bits per code.
For the first coarse search, we ran HNSW for $30,000$ representative vectors ($\boldsymbol{\mu}_1, \dots, \boldsymbol{\mu}_{30,000}$). The distance estimation step was achieved using the proposed SIMD-based search.

\Tref{tbl:large} summarizes the results. Although the accuracy was low, our approach achieved a search in less than 1 ms/query.
To our knowledge, this is the fastest result for the ARM computer with a billion-scale dataset reported in the literature.
Note that our approach provides a highly memory-efficient setting.
The best accuracy is given using a carefully designed approach with data compression and graph-search (e.g., Link\&Code~\cite{cvpr_douze2018}; Recall@1 = 0.668, runtime = 3.5 ms/query, and memory cost = 864 bit/vector).
Compared with Link\&Code, our approach was far less accurate (0.072 vs. 0.668); however, it was fast (0.51 vs. 3.5 ms) and memory-efficient (64 vs. 864 bits/code).

\begin{table}
\centering
\begin{tabular}{@{}llllll@{}} \toprule
& & & & Accuracy & Runtime \\
$n_\mathrm{list}$ & $n_\mathrm{probe}$ & $M$ & $K$ & (Recall@1) & (ms/query) \\ \midrule
30,000 & 1 & 16 & 16 & 0.072 & 0.51 \\
30,000 & 2 & 16 & 16 &  0.082 & 0.83 \\
30,000 & 4 & 16 & 16 &  0.086 & 1.3 \\ \bottomrule
\end{tabular}
\caption{Large-scale evaluation using the Deep1B dataset.}
\label{tbl:large}
\end{table}

\section{conclusion}
We developed a fast SIMD-based acceleration scheme for a 4-bit PQ algorithm using the ARM architecture.
The concept here is to bundle two 128-bit SIMD registers and handle them as a single 256-bit register.
Evaluations showed that our approach accelerates the original PQ by 10x.
We believe that our result is one of the fastest ANN evaluations for ARM to date.
We have already implemented our approach in the Faiss library, and it is freely available for anyone to use.

\textbf{Future work}: The comparison between x86 and ARM machines was not thoroughly performed because it is difficult to compare different architectures in the same experimental setting. Our preliminary benchmark with same-level AWS instances (\texttt{c6i.4xlarge} and \texttt{c6g.4xlarge}) suggested that x86 seems to still be cost-efficient as of 2021. Further evaluation remains as future work.

\textbf{Acknowledgment}: This work was supported by JST, PRESTO Grant Number JPMJPR1936, Japan.

\bibliographystyle{IEEEbib}
\bibliography{refs}

\end{document}